\documentclass[10pt,twocolumn,letterpaper]{article}

\usepackage{cvpr}
\usepackage[]{authblk}
\usepackage{times}
\usepackage{epsfig}
\usepackage{graphicx}
\usepackage{amsmath}
\usepackage{amssymb}
\usepackage{bm}
\usepackage{multirow}
\usepackage[tight,footnotesize]{subfigure}
\usepackage[marginal]{footmisc}
\usepackage{balance}\usepackage{graphicx}
\usepackage{latexsym, amsfonts, amsmath, amsthm, amssymb, mathrsfs, bm}
\usepackage{multirow}
\usepackage{algorithm}
\usepackage{algorithmic}
\usepackage{multirow}
\usepackage[tight,footnotesize]{subfigure}

\usepackage{threeparttable}
\usepackage{amsmath}
\usepackage{dcolumn}
\usepackage{multirow}
\usepackage{booktabs}
\usepackage{mdwlist}
\usepackage[usenames,dvipsnames]{color}
\usepackage{colortbl}
\definecolor{mygray}{gray}{.9}

\cvprfinalcopy 

\def\cvprPaperID{****} 

\begin{document}

\title{Good Practice in CNN Feature Transfer}

\author[1]{Liang Zheng*}
\author[1]{Yali Zhao*}
\author[2]{Jingdong Wang*}
\author[1]{Shengjin Wang}
\author[3]{Qi Tian}
\author{Liang Zheng$^{\dag}$$^{\ddag*}$, Yali Zhao$^{\dag*}$, Shengjin Wang$^{\dag}$, Jingdong Wang$^{\S}$, Qi Tian$^{\ddag}$  \\$^{\ddag}$University of Texas at San Antonio \qquad  $^{\dag}$Tsinghua University \qquad $^{\S} $Microsoft Research }
\maketitle
\thispagestyle{empty}

\begin{abstract}
The \footnote{\noindent{* Two authors contribute equally to this work.}} objective of this paper is the effective transfer of the Convolutional Neural Network (CNN) feature in image search and classification. Systematically, we study three facts in CNN transfer. 1) We demonstrate the advantage of using images with a properly large size as input to CNN instead of the conventionally resized one. 2) We benchmark the performance of different CNN layers improved by average/max pooling on the feature maps. Our observation suggests that the Conv5 feature yields very competitive accuracy under such pooling step. 3) We find that the simple combination of pooled features extracted across various CNN layers is effective in collecting evidences from both low and high level descriptors. Following these good practices, we are capable of improving the state of the art on a number of benchmarks to a large margin.
\end{abstract}

\section{Introduction}
The Convolutional Neural Network (CNN) has been record-leading in a number of vision tasks, \emph{e.g.,} image recognition \cite{krizhevsky2012imagenet, simonyan2014very, szegedy2014going} and segmentation \cite{long2014fully}, object detection \cite{he2014spatial, girshick2014rich}, instance retrieval \cite{sharif2015baseline, razavian2014cnn, babenko2014neural}, \emph{etc}. CNN transfer is a common practice in the vision community due to the expense in collecting sufficient amount of training data. A typical example consists in instance-level image search, in which it is infeasible to collect training data considering the wide variety in query content. Another example includes the fine-grained classification: experts are needed for class annotation, prohibiting the access to the large amount of training data. Considering these challenges, this paper is devoted to the effective usage of pre-trained CNN models in image search and classification.

Our study is motivated by three aspects. First, images used as input to CNN in most current works are resized to a fixed size, \emph{e.g.,} $227\times227$ for AlexNet \cite{krizhevsky2012imagenet}. It ensures that the CNN output is a $1\times1\times N$ vector, where $N$ is the number of blobs in the fully connected layer. And yet, this process may suffer from information loss during image down-sampling. In previous works, Simonyan \emph{et al.} \cite{simonyan2014very} merge classification results from multi-scale image inputs on the ILSVRC'12 validation set \cite{ILSVRC15}. In this work, we initialize a comprehensive study of this issue on a number of transfer datasets.

Second, features from the fully connected (FC) layers are mostly used for transfer \cite{he2014spatial, girshick2014rich, sharif2015baseline, razavian2014cnn, babenko2014neural}.  These global features are trained towards assigning category labels to images and to some extent invariant to illumination, rotation, \emph{etc}. The FC features, however, suffer from the lack of description of local patterns, which is especially critical when occlusion or truncation exists \cite{razavian2014cnn, babenko2014neural}. With respect to the sensitivity to local stimulus, CNN features from bottom or intermediate layers have shown promises \cite{he2014spatial, ng2015exploiting}. These discriminatively trained convolutional kernels respond to specific visual patterns that evolve from bottom to top layers. While capturing local activations, the intermediate features are less invariant to image translations. The findings in this paper suggest that under proper pooling steps, the performance lower-level features undergoes impressive improvement.

Third, it is known that the top layers in CNN are sensitive to semantics, while intermediate layers are specific to low-level patterns, such as oriented bars. In transfer problems, it is not trivial to predict which layers are superior. For example, in fine grained classification, finer details in images are more important than high-level semantics; in generic classification, it may well be the case that top layers work better. Previous works \cite{he2014spatial, SPM, gong2014multi} typically perform multi-scale analysis through image-level partitions. This paper instead inputs a single-scale image, obtains multi-layer CNN features through one feed-forward step, and fuses the features to further improve recognition accuracy.

\begin{figure*}
  \centering
  \includegraphics[width=6.8in]{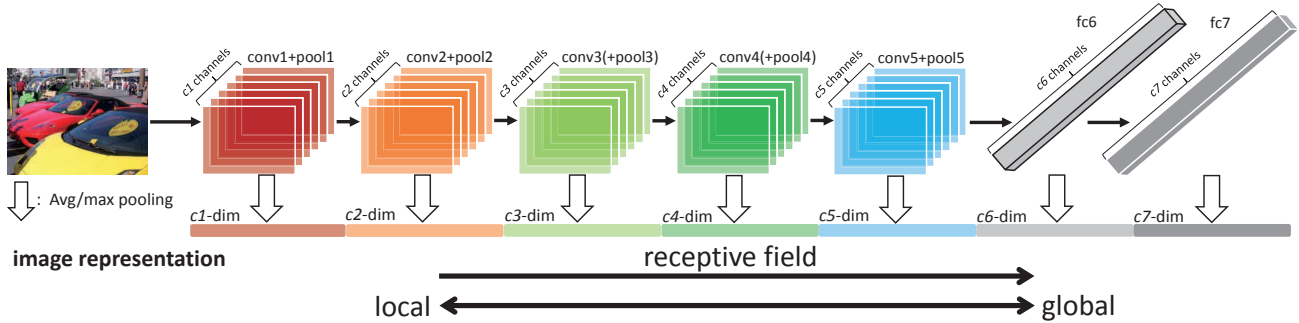}\\
  \caption{Feature extraction scheme. An input image of arbitrary size is fed forward in the CNN network. Feature maps in layer $k$ consist of $c_k$ channels. Then, average or max pooling is employed to generate a $c_k$-dim feature vector. We use for image classification and search either the pooled feature from a single layer or the concatenated pooled feature from all layers. }\label{fig:method}
\end{figure*}

Starting from the above issue, this paper provides some good practices that are suggested during CNN feature transfer. We share in this paper three findings contributing to the improvement over state-of-the-arts on a serial of benchmarks.
\begin{itemize}
\item Evidence accumulates that using images with larger sizes as input to CNN is a better choice and yields consistent improvement during CNN transfer.
\item We observe that average/max pooling of features from intermediate layers is effective in improving invariance to image translations. Specifically, the pooled Conv5 feature, with much lower dimensionality, is shown to yield competitive accuracy with the FC features.
\item We find it beneficial to fuse pooled features from multiple layers. Along the feature hierarchy, the receptive field increases, and features tend to capture more global cures. By fusing features with various receptive fields, information of multiple scales (local and global) is effectively combined.
\end{itemize}

Following the above-mentioned practices, we report results on 3 image search and 7 image classification benchmarks, and present significant improvement over the state-of-the-art methods. The pipeline of the feature extraction routine is illustrated in Fig. \ref{fig:method}. Throughout this paper, unless specified, we use term ``Conv$X$'' to refer to layer ``Pool$X$ (Conv$X$ + max pooling)'', to avoid confusion with our pooling step.

The remainder of this paper is organized as follows. In Section \ref{sec:related_work}, we will briefly review related literature. Section \ref{sec:method} provides detailed description of the proposed representation. Section \ref{sec:experiment} summarizes and presents the experimental results and conclusions are drawn in Section \ref{sec:conclusion}.

\section{Related Works}\label{sec:related_work}
We will provide a brief literature review from several closely related aspects, \emph{i.e.,} comparison of CNN features from different layers, combining features across multiple scales, and image search/classification using CNN.

\textbf{Comparison of CNN features from different layers.} In most cases, features from the fully connected (FC) layers are preferred, with sporadic reports on intermediate features. For the former, good examples include ``Regions with Convolutional Neural Network Features'' (R-CNN) \cite{girshick2014rich}, CNN baselines for recognition \cite{razavian2014cnn}, Neural Codes \cite{babenko2014neural}, \emph{etc}. The prevalent usage of FC features is mainly attributed to its strong generalization and semantics-descriptive ability. Regarding intermediate features, on the other hand, results of He \emph{et al.} \cite{he2014spatial} on the Caltech-101 dataset \cite{fei2006one} suggest that the Conv5 feature is superior if Spatial Pyramid Pooling (SPP) is used, and is inferior to FC6 if no pooling step is taken. Xu \emph{et al.} \cite{xu2014discriminative} find that the VLAD \cite{jegou2011product} encoded Conv5 features produce higher accuracy on the MEDTest14 dataset in event detection. In \cite{ng2015exploiting}, Ng \emph{et al.} observe that better performance in image search appears with intermediate layers of GoogLeNet \cite{szegedy2014going} and VGGNet \cite{simonyan2014very} when VLAD encoding is used. This paper demonstrates the competitiveness of Conv5 with FC features using simple pooling techniques. In a contemporary work, Mousavian \emph{et al.} \cite{mousavian2015deep} draw similar insights in image search. Our works is carried out independently and provides a comprehensive evaluation of intermediate features on both image search and classification.

\begin{figure*}
  \centering
  \includegraphics[width=6.8in]{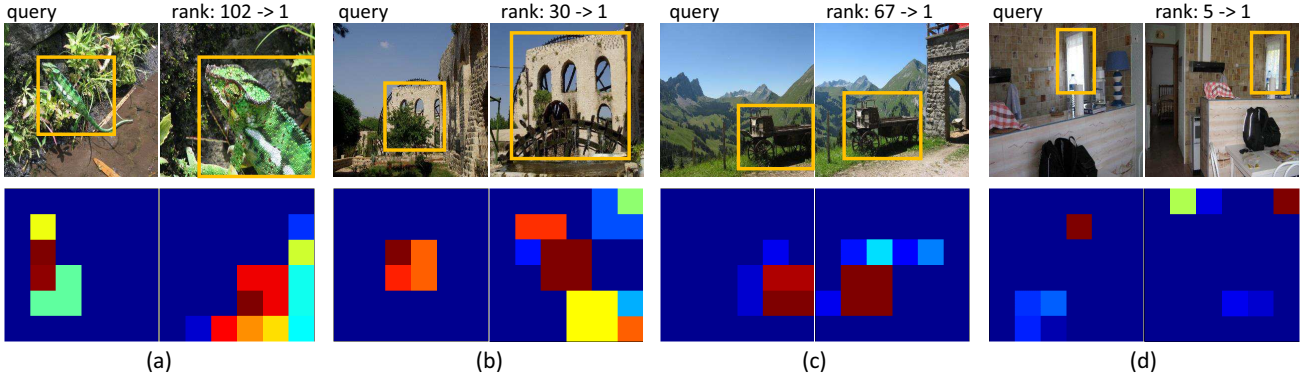}\\
  \caption{Illustration of the advantage of Conv5 feature in image search. Four image pairs are shown. To the left of each pair is the query image, and to the right the relevant database image. The second row depicts the feature maps from a certain channel of the Conv5 feature of the corresponding images. For each relevant image, its ranks obtained by both the FC6 and Conv5 features (after average pooling) are shown. Clearly, the pooled Conv5 feature captures mid-level cues and improves search accuracy. }\label{fig:conv5}
\end{figure*}
\textbf{Combining features across multiple scales.} The integration of multi-scale cues has been proven to bring decent improvements. A well-known example consists in Spatial Pyramid Matching (SPM) \cite{SPM}, which pools Bag-of-Words vectors from multiple image scales to form a long vector. In works associated with CNN, Gong \emph{et al} \cite{gong2014multi} pool CNN features extracted from multi-scale image patches for image search and classification. Yoo \emph{et al} \cite{yoo2015multi} propose to encode dense activations of the FC layers with Fisher Vector \cite{perronnin2007fisher} from multiple image scales for image classification. In image segmentation, Farabet \emph{et al} \cite{farabet2013learning} concatenate CNN features from the same CNN layer but from multiple scales of the image, so the features have similar invariance. The closest work to ours is the ``Hypercolumn'' \cite{hariharan2014hypercolumns}, in which Hariharan \emph{et al} address object segmentation and pose estimation by resizing convolutional maps in each layer and concatenating them on each pixel position. This paper instead focuses on holistic image recognition by fusing pooled vectors from various layers in the CNN structure, and demonstrates consistent improvement on the benchmarks.

\textbf{Image search/classification using CNN.}  In \textbf{image search}, CNN can be used as global \cite{babenko2014neural, xie2015image, razavian2014cnn}, regional \cite{gong2014multi, sharif2015baseline}, or local features \cite{ng2015exploiting}. Basically in image search, CNN features are required to be memory efficient, so encoding or hashing schemes are beneficial. Compared with traditional search framework \cite{AKM} based on the SIFT descriptor \cite{SIFT2} and the inverted index, CNN feature is more flexible and yields superior results \cite{sharif2015baseline}. One problem of CNN feature is the lack of invariance to rotation, occlusion, truncation. So its usage as local feature \cite{ng2015exploiting} or the inclusion of rotated patches \cite{xie2015image} are good choices against these limitations. On the other hand, \textbf{image classification} has been greatly advanced by the introduction of CNN \cite{krizhevsky2012imagenet, razavian2014cnn}. CNN features are mainly used as global \cite{krizhevsky2012imagenet} or part \cite{zhang2014part} descriptors, and are shown to outperform classic hand-crafted ones in both transfer and non-transfer classification tasks. While baseline results with FC features have been reported on small- or medium-sized datasets \cite{razavian2014cnn}, a detailed evaluation of different layers as well as their combination is still lacking. This paper aims at filling this gap by conducting comprehensive empirical studies.

\section{Method}\label{sec:method}
We will describe the pooling and fusion steps (see Fig. \ref{fig:method}) in Section \ref{sec:pooling_step} and Section \ref{sec:fusion}, respectively. The proposed image resizing protocol will be presented in Section \ref{sec:role_size}.
\subsection{Pooling}\label{sec:pooling_step}
Given a pre-trained CNN network $\mathcal{C}$, there are $K$ convolutional layers, denoted as $L^1, L^2, ..., L^K$. Here, the fully connected layers are summarized as convolutional layers as well, because full connection is a special case of convolution \cite{simonyan2014very}. Also, each convolutional layer may include a max pooling sub-layer. In AlexNet, there are 8 convolutional layers. In VGGNet, we use the 19-layer model, in which there are 8 major convolutional layers. We take the last sub-layer as a representative of the major layer. For example, among sub-layers $5-1, 5-2, 5-3$ and $5-4$ in layer 5, we use $5-4$ on behalf.

Given an image $d$, we denote the feature maps of layer $k$ as $f^k (k = 1,...,K)$ . Assume that $f^k$  takes the size of $w_d^k \times h_d^k \times c^k$, where $w_d^k$ and $h_d^k$ are the width and height of each channel, respectively, and $c^k$ denotes the number of channels (or convolutional kernels) of layer $L_k$. Note that, for input images with different sizes, the size of the convolutional maps can be different. Then, we exert average or max pooling steps on the maps, \emph{i.e.,}
\begin{equation}\label{eq:avg_pooling}
 p_{avg}^k(i) = \frac{1}{w \times h} \sum{f^k(\cdot, \cdot, i)}, i=1,2,...c^k,
\end{equation}
\begin{equation}\label{eq:avg_pooling}
  p_{max}^k(i) = \max{f^k(\cdot, \cdot, i)}, i=1,2,...c^k,
\end{equation}
where $p_{avg}^k$ and $p_{max}^k$ are both $1\times1\times c^k$ dimensional, representing the output of average and max pooling, respectively. In VGGNet, for example, the result of average/max pooling on Conv5 is of dimension $1\times1\times512$.\\

\noindent\textbf{Discussion.}
  By pooling intermediate convolutional maps, there are two main advantages. First, using the intermediate features, the local structures (instead of the global cues by FC features that are mostly used) are paid more attention to. This is because the convolutional filters are sensitive to specific visual patterns, ranging from low-level bars to mid-level parts. Second, by pooling, the resulting vectors have higher invariance to translation, occlusion, and truncation of the local stimulus, which greatly improves the effectiveness of the intermediate features. An additional advantage of pooling is the computational efficiency brought about by low-dimensional feature vectors (512-dim for VGGNet Conv5 feature).

Figure \ref{fig:conv5} illustrates some examples where conv5 feature (followed with average pooling) captures common local structures in relevant image pairs which are largely lost in the FC6 representation. For example, Fig. \ref{fig:conv5}(c) and Fig. \ref{fig:conv5}(d) each show two images containing a similar pattern at distinct positions. The convolutional maps (second row in Fig. \ref{fig:conv5}) that respond to such patterns are very different due to the intensive image variations. In this example, average pooling alleviates the influence of translation variance and improves the rank of the relevant image.  In another example, when truncation (Fig. \ref{fig:conv5}(a)) exists, the top-layer feature (FC6) is less effective because the high-level information is to some extent blocked. In this case, low-level features may be of great value by detecting common local patterns between the two images. Then, average pooling improves truncation invariance by capturing the local similarities.

\subsection{Fusion}\label{sec:fusion}
   We combine features from different layers, \emph{i.e.,} from low to high levels, and from small to large receptive fields.

\textbf{For image search}, simply concatenating the multi-level features probably does not work because it assumes the same weight for all features, which is clearly undesirable without a learning process. To make feature weight adaptive to the query, this paper adopts the state-of-the-art late fusion strategy proposed in \cite{zheng2015query}. In a nutshell, this method adaptively assigns different weights to the scores of different features through offline reference collection and online reference selection steps. With this method, we dynamically fuse features across various CNN layers.

\textbf{For image classification}, simple feature concatenation is employed. Given an input image, we extract features from all seven layers, \emph{i.e.,} Conv1, Conv2, ..., Conv5, FC6, and FC7. Note that the input image may of arbitrary sizes. We denote the seven features as $f^i (i=1,2,...,7)$, and after pooling (max or average), as $p^i (i = 1, 2, ..., 7)$. Then, the visual representation of the input image is written as $(p^1, p^2, ..., p^5, p^6, p^7)^T$. For AlexNet, the dimension of the final representation is $96+256+384+384+256+4,096+4,096 = 9,568$; For VGGNet, the feature dimension is calculated as $64+128+256+512+512+4,096+4,096 = 9,664$. Figure \ref{fig:method} provides an illustration of feature extraction.\\

\noindent\textbf{Discussion.} The CNN features from bottom layers have very high dimensionality. For example, in AlexNet, the dimensions of the raw features from pool1 and pool2 are 69,984 and 43,264, respectively, for an image of size $227 \times 227$. The feature dimension will be even higher if larger images are used as input. The direct usage of such high-dimensional vectors is prohibitive from efficient search or classifier training. This is probably one reason why previous works \cite{farabet2013learning, gong2014multi, yoo2015multi} typically use images of multiple scales and concatenate/pool CNN features from the same FC layer. In our work, we demonstrate a promising way of using much shorter feature representations and fuse features encoding increasing semantic levels.

\subsection{The Role of Image Size} \label{sec:role_size}
In conventional cases, images are resized so that all images have the same size. In AlexNet, for example, during CNN training, all images are resized to $227\times227$ before being fed into the network. During testing, images are typically resized to $227\times227$ for feature extraction and classification.

In transfer tasks, however, the pre-trained CNN models may not fit well the distribution of an unseen dataset. Under such cases, the simple image resizing does not have a solid back up. Resizing large images into $227\times227$ (or $224\times224$ in VGGNet) may suffer from substantial information loss and object distortion. This problem is not trivial, as in object retrieval, the object-of-interest may take up only a small region in the target image, and in a larger image, details can be more clearly observed. Also, object distortion may compromise image matching between the query and database images. In recognition, keeping aspect ratio of an image will also help preserve the shape of objects/scene, thus being beneficial for accurate classification.

\begin{figure*}
  \centering
  \includegraphics[width=6.8in]{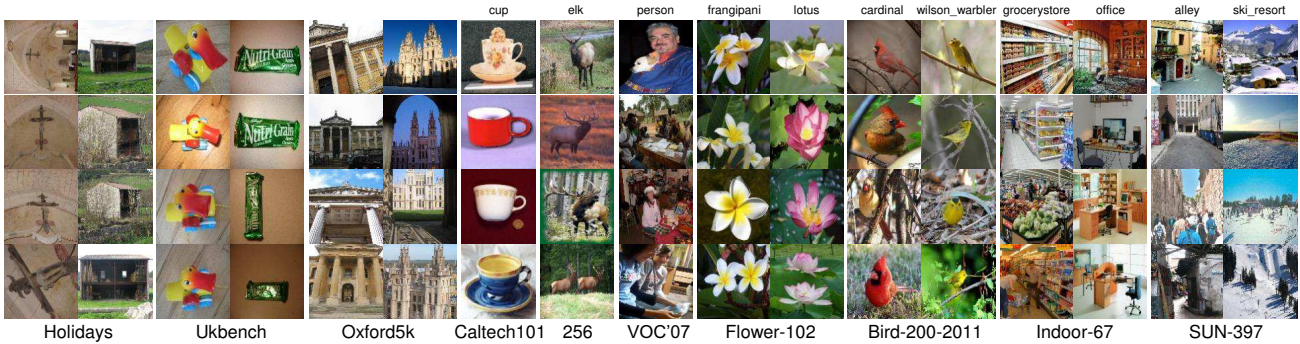}\\
  \caption{Sample images of the experimental datasets. For each dataset, one or two groups of images are listed that depict relevant content (image search) or belong the same class (image classification). The class labels are also shown above the columns of classification datasets. }\label{fig:sample_images}
\end{figure*}
\setlength{\tabcolsep}{6.6pt}
\begin{table*}[!t]
\renewcommand{\arraystretch}{1.0}
\begin{center}
\caption{Average side lengths of the 10 datasets.}
\begin{tabular}{|l|ccccccc|ccc|}
\hline
\multirow{1}{*}{Datasets} &
 Bird & Flower & Indoor & SUN & Cal-101 & Cal-256 & VOC'07 &Holidays & Ukbench & Oxford\\
\hline
\hline
Long side & 490.8 & 664.0 & 521.3 & 1,002.8 & 319.7 & 398.6 &496.4 & 1,024.0  & 640.0 & 1,024.0\\
\hline
Short side & 364.3 & 500.0  &  399.1   & 733.2 &227.0  & 296.9 & 358.2 & 768.0 &480.0 & 746.0\\
\hline
\end{tabular}
\label{table:image_size}
\end{center}
\end{table*}

Consequently, in this paper, relatively large images (compared with $224\times224$) are taken as input. Specifically, given a dataset, we calculate the average image height and width of the training set. Then, the larger value between the average height and width is taken as the long side of all training and testing images. For example, if the average image size of the training set is $400\times300$, we will accordingly resize all images into a longer size of 400 pixels, and keep the aspect ratio. In image search, we calculate the average side lengths from the database images. In this paper, the image size calculated in this manner is termed \textbf{scale 1.0}. The other scales are defined as the ratio between their long side to that of scale 1.0. For example, if a long size of 400 pixels is defined as scale 1.0, then scaling it to 300 pixels yields a scale of 0.75. In Section \ref{sec:experiment}, several image scales will be tested to show the advantage of this protocol.

\makeatother
\begin{figure*} [t]
\centering
\subfigure[Bird-200-2011]{\label{fig:six_bird}%
\includegraphics[width=2.0in]{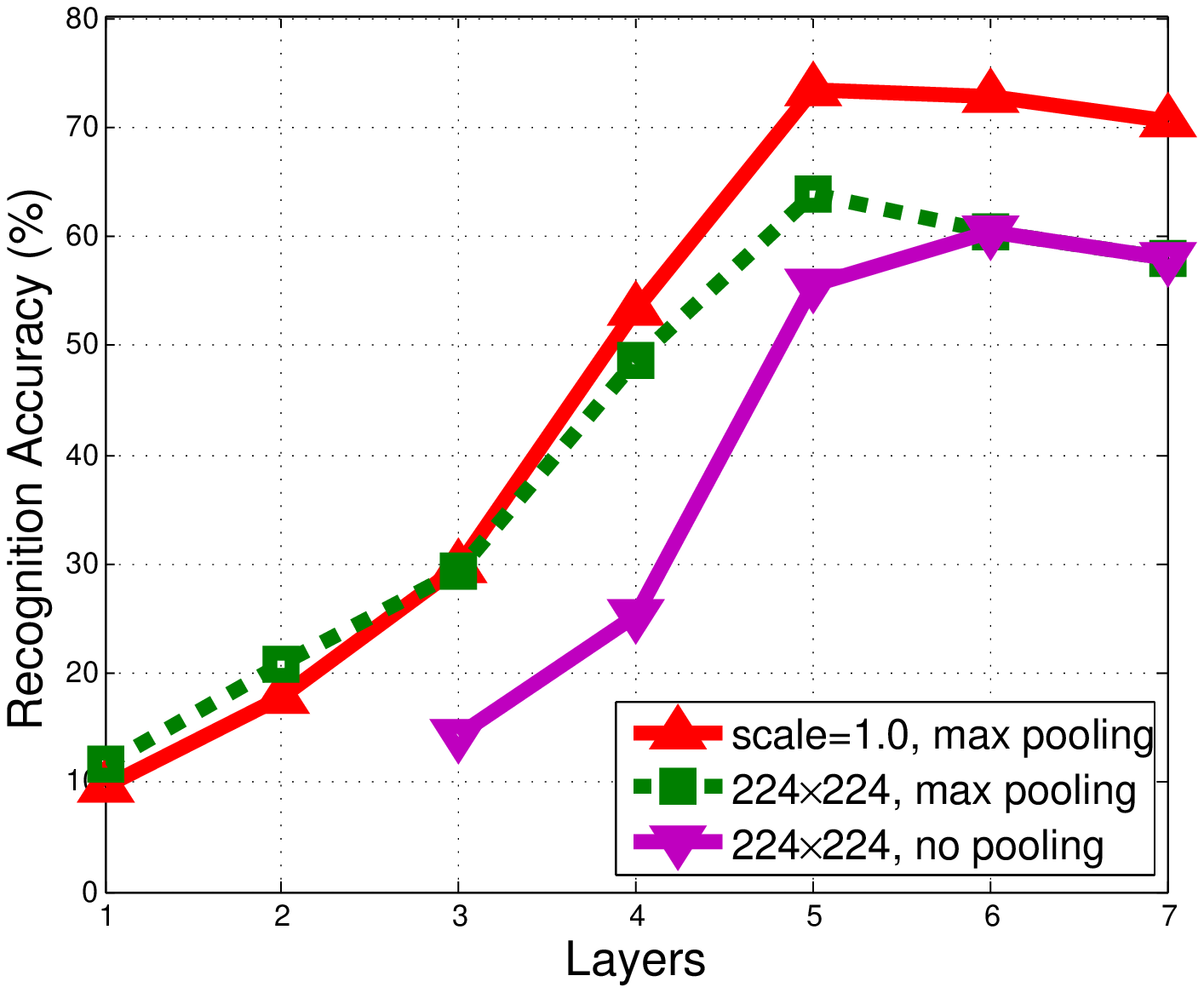}}
\hspace{0.18in}
 \subfigure[Flower-102]{\label{fig:six_flower}%
\includegraphics[width=2.0in]{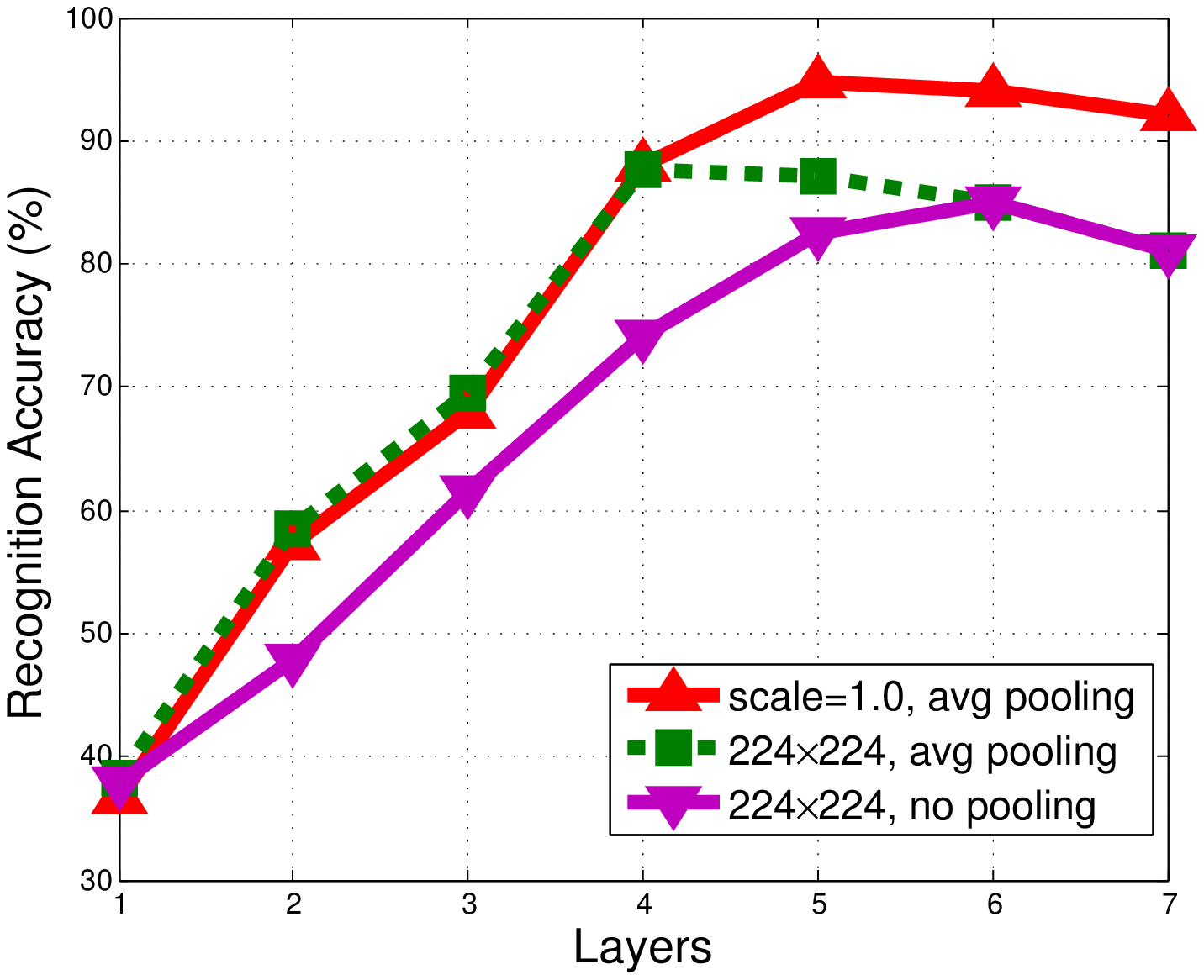}}
\hspace{0.18in}
 \subfigure[Indoor-67]{\label{fig:six_indoor}%
\includegraphics[width=2.0in]{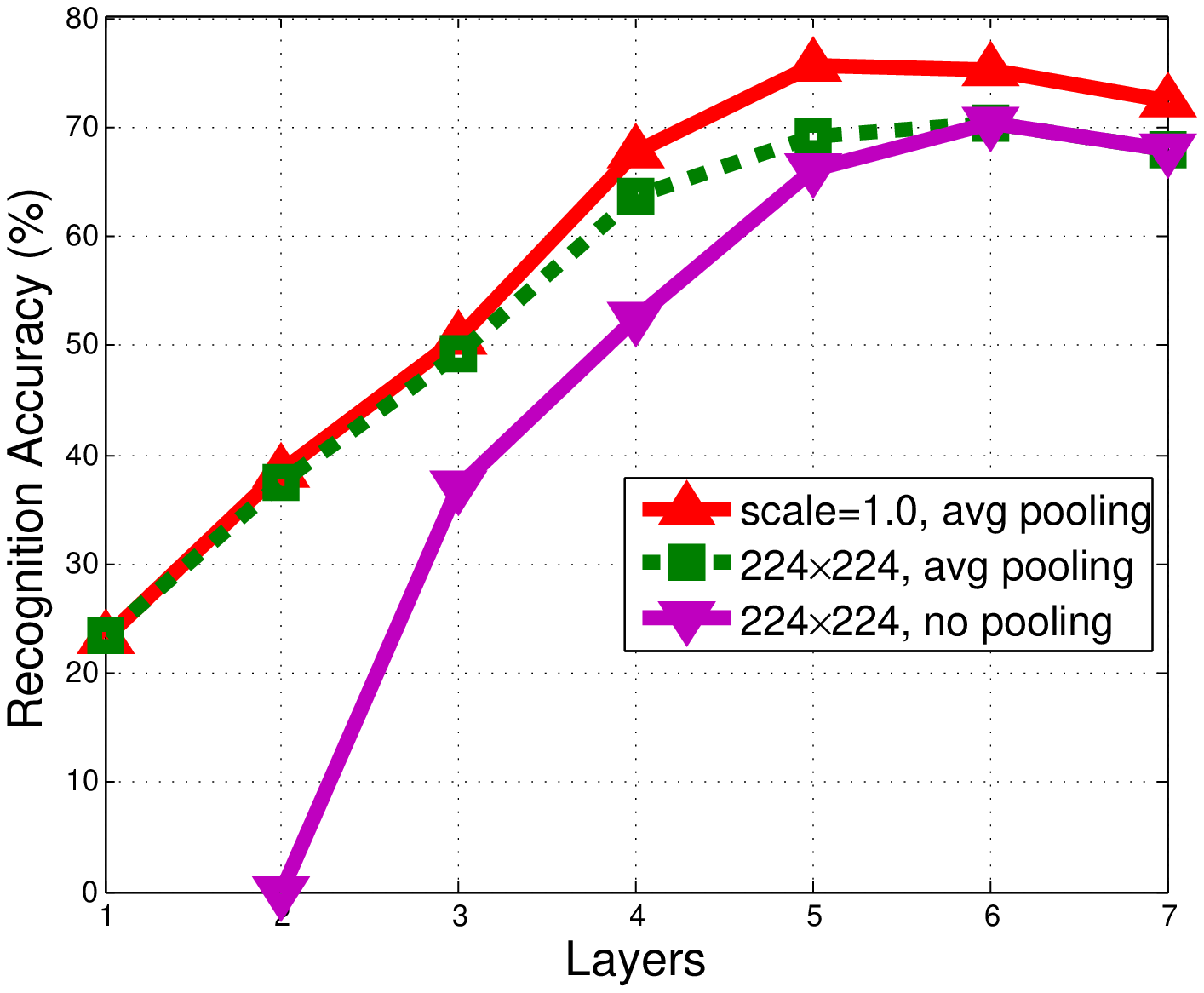}}
\subfigure[Holidays]{\label{fig:six_holidays}%
\includegraphics[width=2.0in]{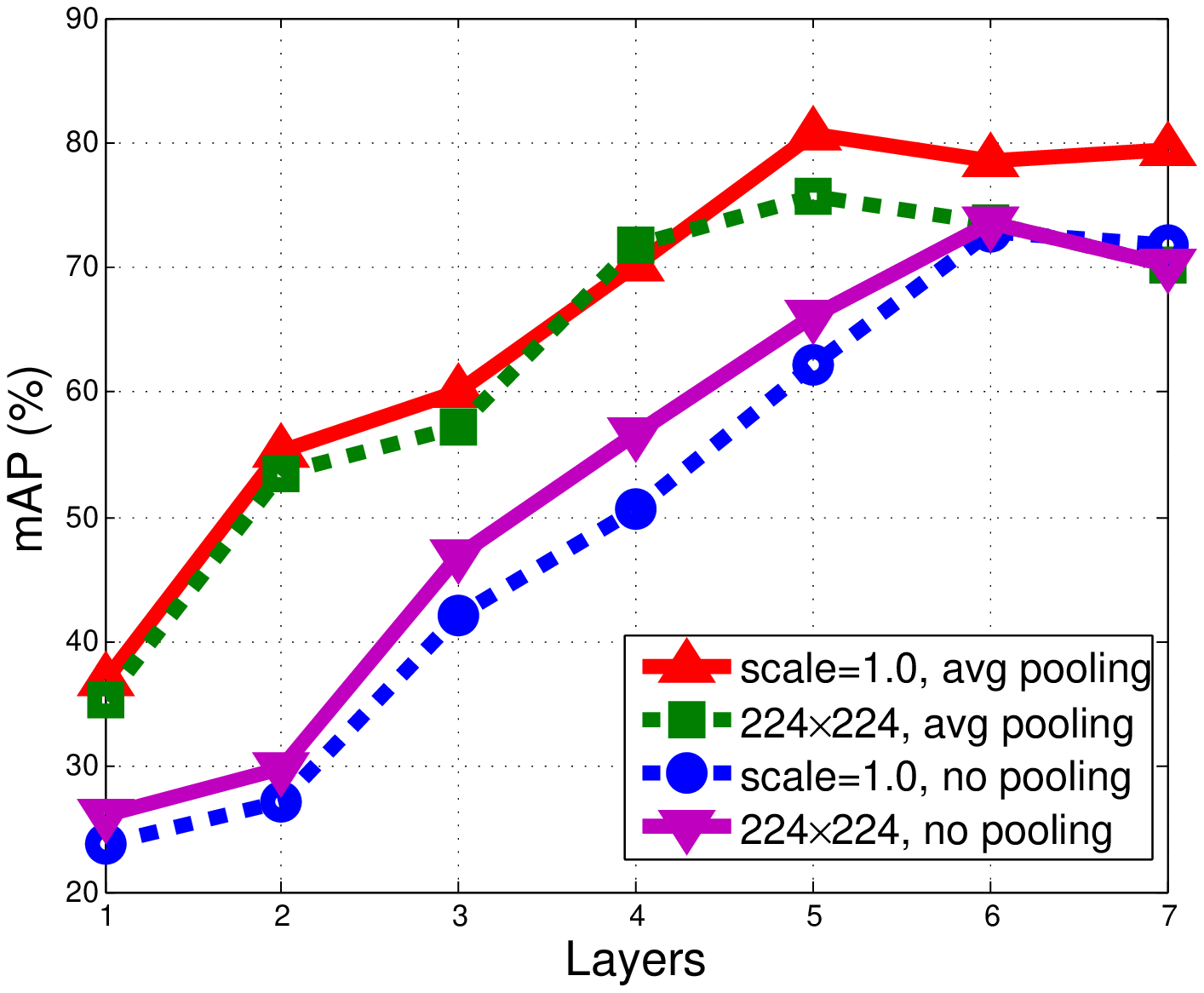}}
\hspace{0.18in}
 \subfigure[Ukbench]{\label{fig:six_ukbench}%
\includegraphics[width=2.0in]{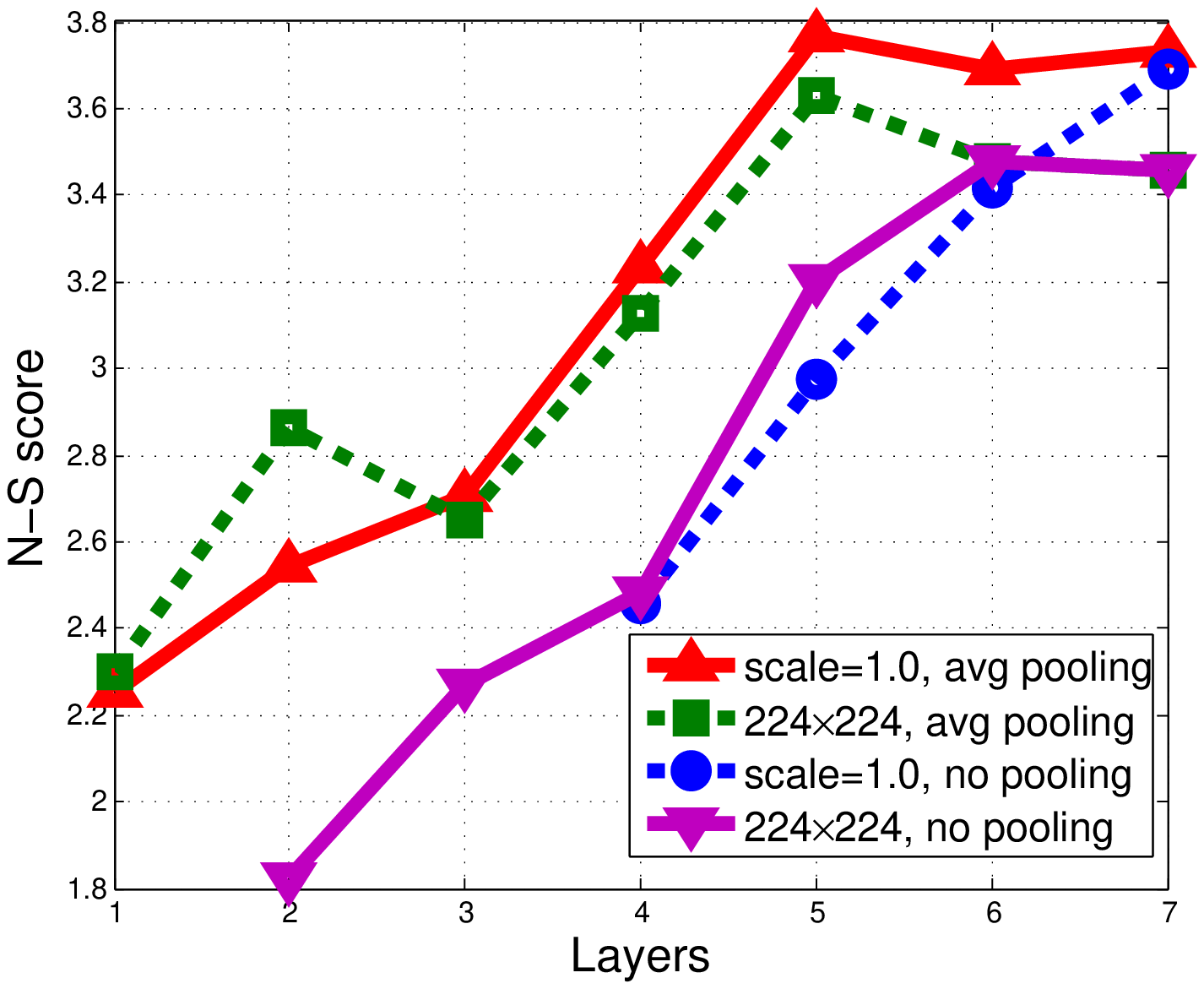}}
\hspace{0.18in}
 \subfigure[Oxford5k]{\label{fig:six_oxford}%
\includegraphics[width=2.0in]{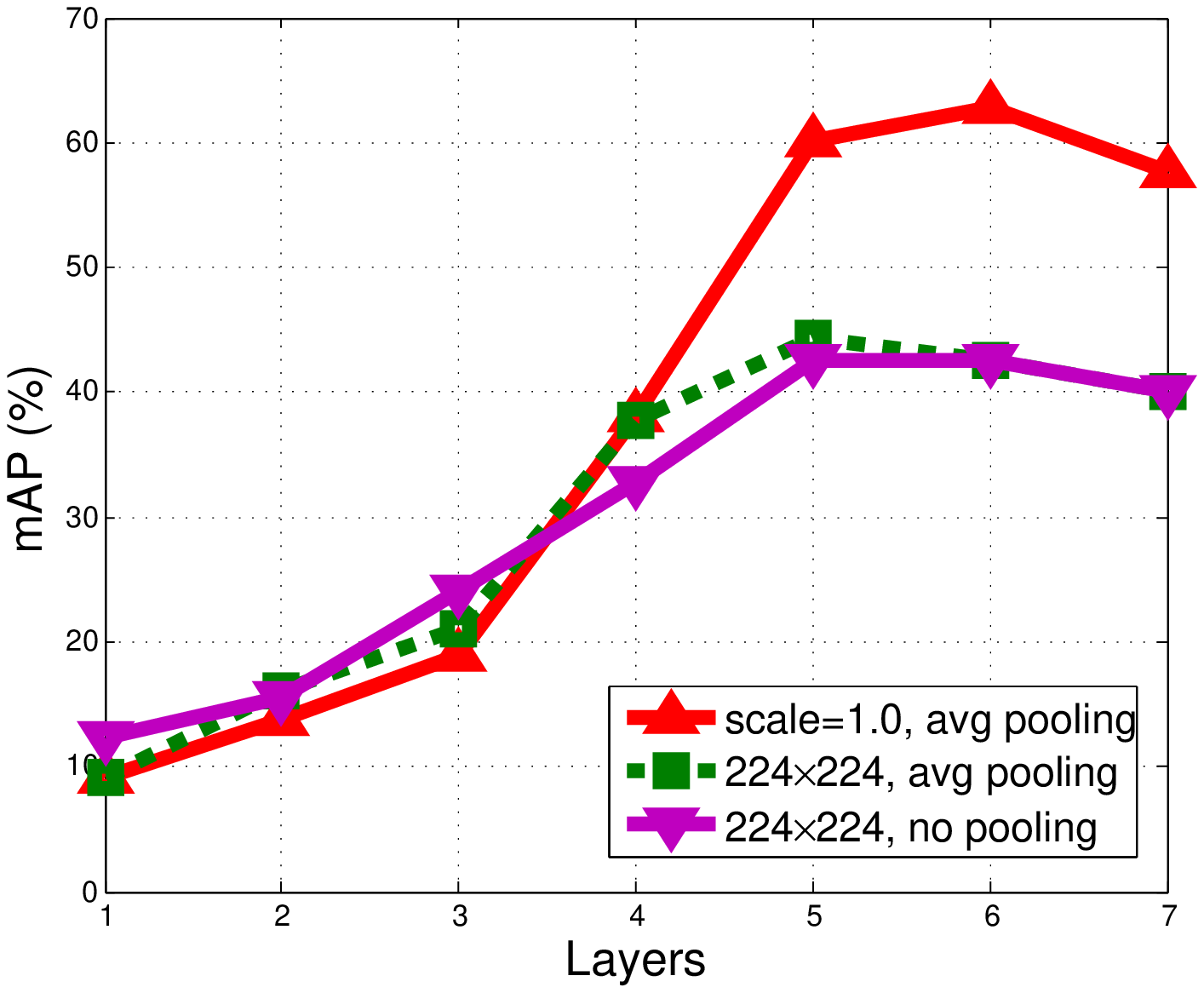}}
\caption{Image search and classification accuracy on six benchmarks. We plot the accuracy against features extracted from different CNN layers (VGGNet). For each dataset, we compare two image sizes, \emph{i.e.,} $224\times224$ and scale 1.0 (see Section \ref{sec:role_size}); also, we compare the cases when no pooling is used and when max or avg pooling is used. }
\label{fig:six_datasets}
\end {figure*}
\section{Experiments}\label{sec:experiment}
\subsection{Datasets}\label{sec:dataset}
In image search, we use three datasets, \emph{i.e.,} \textbf{Holidays} \cite{jegou2010improving}, \textbf{Ukbench} \cite{HKM}, and \textbf{Oxford5k} \cite{AKM}. The Holidays dataset consists of 1,491 scene images among which 500 are selected as queries. The Ukbench dataset contains 10,200 images divided into 2,550 groups, each representing a unique object or scene. All images are taken as queries in turns, and there are 4 ground truths for each query. The Oxford5k dataset contains 5,063 images, crawled by names of the architectures in Oxford. There are 55 query buildings well-defined by hand-drawn bounding boxes. Mean Average Precision (mAP) is taken as evaluation metric for Holidays and Oxford5k datasets, while N-S score (average number of relevant images in top-4 ranked images) is used for the Ukbench dataset.

In image classification, 7 datasets are tested. For generic classification, we use \textbf{Caltech-101} \cite{fei2006one}, and \textbf{Caltech-256} \cite{griffin2007caltech} and \textbf{PASCAL VOC'07} \cite{Everingham10} datasets. 30 and 60 images per category are randomly selected for training on Caltech-101 and -256, respectively. For VOC'07, we use the standard train/test split, and calculate the mean Average Precision (mAP) over 20 classes. Then, for scene classification, we evaluate on \textbf{Indoor-67 } \cite{quattoni2009recognizing} and \textbf{SUN-397} \cite{xiao2010sun} datasets, in which 80 and 50 training images per category are selected, respectively. For fine-grained classification, Oxford \textbf{Flower-102} \cite{nilsback2008automated} and  Caltech-UCSD \textbf{Bird-200-2011} \cite{wah2011caltech} datasets are used. 20 and 30 training images per category are randomly chosen, respectively. For all the datasets except VOC'07, we repeat the random partitioning of training/test images for 10 times, and the averaged classification accuracy is reported. Sample images of the 10 datasets are shown in Fig. \ref{fig:sample_images}. The average side lengths of the datasets are presented in Table \ref{table:image_size}.

\subsection{Implementation Details}\label{sec:details}
We mainly report the performance of VGGNet \cite{simonyan2014very} for brevity and due to its state-of-the-art accuracy in ILSVRC'14 \cite{ILSVRC15}. We also present the performance of AlexNet \cite{krizhevsky2012imagenet} in our final system. Both CNNs are pre-trained on the ILSVRC'12 dataset \cite{deng2009imagenet}. The VGGNet has 19 layers which can be divided into 8 major convolutional layers (including 3 fully connected layers), and the AlexNet is composed of 8 convolutional layers (including 3 fully connected layers). Both models are available online \cite{jia2014caffe}.

For pooled features from each layer, we use the square root normalization introduced in \cite{root_sift}. Namely, we exert a square root operator on each dimension, and then $l_2$-normalize the vector. Then, after concatenating features from all layers, another $l_2$-normalization is employed. For VGGNet and AlexNet, the dimensions of the pooled Conv5 features are 256 and 512, respectively. For VGGNet, the five convolutional layers each have 2, 2, 4, 4, and 4 sub-layers, respectively. We take the last sub-layer as the representative of the convolutional layer (see Section \ref{sec:pooling_step}).

\subsection{Evaluation of Individual Steps}\label{sec:Individual}
\textbf{Performance of different CNN layers.}
Features from different CNN layers vary in their receptive field sizes and semantic levels. To evaluate their impact on recognition and search accuracy, we test the 7 layers on six datasets, \emph{i.e.,} Bird-200, Flower-102, Indoor-67, Holidays, Ukbench, and Oxford5k, and results are presented in Fig. \ref{fig:six_datasets} and Table \ref{table:various_approaches}.

Findings are consistent across the six datasets: features from the bottom layers are generally inferior to those from the top layers. This is expected because bottom-layer filters are sensitive to low-level visual patterns which are generally considered not discriminative enough. These filters to some extent resemble the visual words in the Bag-of-Words model with the SIFT descriptor \cite{SIFT2}. On the Bird-200-2011 dataset, for example, the recognition accuracy of the Conv1 feature with max pooling is only 11.62\% by conventional $224\times224$ resizing. From this observation, it is desirable that further discriminative cues be incorporated in the bottom features, \emph{e.g.,} spatial constraints \cite{GVP}, or multi-feature fusion \cite{zheng2015query}.

\setlength{\tabcolsep}{5.8pt}
\begin{table*}[!t]
\renewcommand{\arraystretch}{1.0}
\begin{center}
\caption{Results on 10 benchmarks \emph{w.r.t} different CNN layers with average/max (a/m) pooling. Max pooling is used on the Bird dataset.}
\begin{tabular}{|l|ccccccc|ccc|}
\hline
\multirow{1}{*}{Datasets} &
 Bird & Flower & Indoor & SUN & Cal-101 & Cal-256 & VOC'07&  Holidays & Ukbench & Oxford\\
\hline
\hline
conv4+a/m pool. & 53.20 & 88.01 &  67.81 & 50.71 & 80.44 & 63.86 & 67.55 & 70.25 & 3.23 & 38.10\\
\rowcolor{mygray}
conv5+a/m pool. & \textbf{73.40} &  \textbf{94.73}  &    \textbf{75.67}   & \textbf{58.88}  & 91.07 & 83.29 & 81.78 &\textbf{80.71} & \textbf{3.77} & 60.18\\
FC6+a/m pool. & 72.78 &  94.07  &  75.32     &  57.76 & 92.24& 84.20  &82.31 & 78.46 & 3.69 & 62.77\\
FC7+a/m pool. & 70.64 &  92.05  &   71.4    & 58.31  & 89.28 & 83.82& 82.57 & 79.43 &  3.73& 57.63\\
\hline
\rowcolor{mygray}
All layers & \textbf{76.35} & \textbf{95.62} & \textbf{78.42} &  \textbf{63.71} &  \textbf{92.31} & \textbf{85.99} & \textbf{83.66}& \textbf{84.2} & \textbf{3.75} & \textbf{71.30}\\
\hline

\end{tabular}
\label{table:various_approaches}
\end{center}
\end{table*}

There is one noticeable observation that \textbf{the ``Conv5 + a/m pooling'' feature yields very competitive performance to the FC6 and FC7 features.} For image search, ``Conv5 + avg pooling'' improves search accuracy by +1.28\% and +0.04 over FC6 and FC7 on Holidays and Ukbench, in mAP and N-S score, respectively. For image classification, Conv5 feature is superior in 4 out of 7 datasets, \emph{i.e.,} on fine-grained and scene classification tasks. In fact, fine-grained and scene classification are representative transfer tasks in which the predefined classes are more distant from those defined in ILSVRC \cite{deng2009imagenet}. Moreover, in both tasks, local or mid-level elements are useful cues to discriminate between classes. In such scenarios, the Conv5 feature is advantageous in that it encodes mid-level activations which further acquire translation/occlusion invariance through the pooling step. Experiment confirms our assumption: classification accuracy improves by +0.62\%, +0.66\%, +0.35\%, and +1.12\% on Bird, Flower, Indoor, and SUN, respectively.

For generic recognition as the case in Caltech-101, Caltech-256, and PASCAL VOC'07, FC6 outperforms conv5 feature due to two reasons. First, objects in the first two datasets are well-positioned in the center of the image: translation and occlusion are not severe. Second, the VGGNet is trained on generic ILSVRC classification dataset: it is less a transfer problem for generic classification task.

\makeatother
\begin{figure*} [t]
\centering
\subfigure[conv5+avg pooling]{\label{fig:size_conv5}%
\includegraphics[width=2.0in]{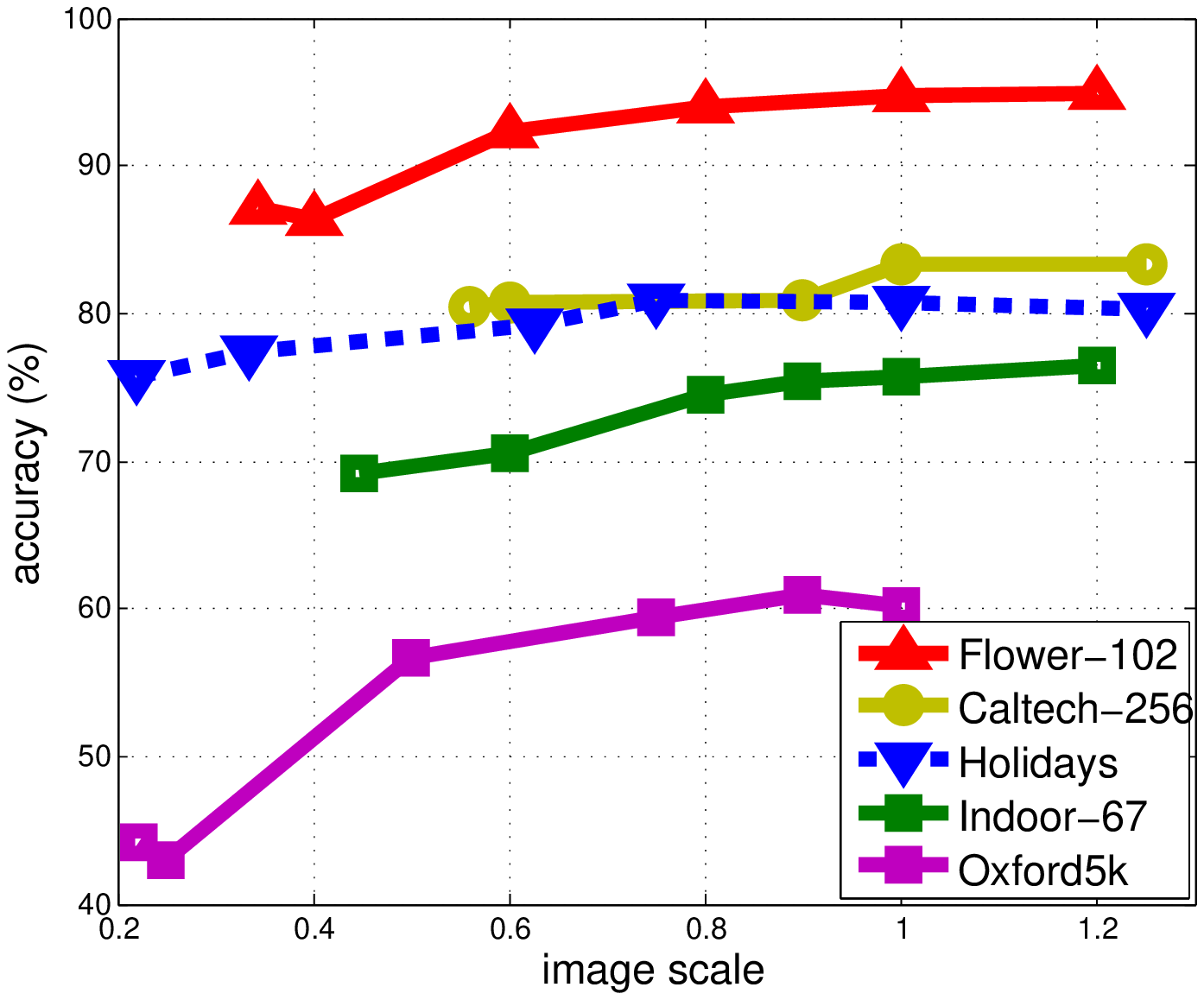}}
\hspace{0.18in}
 \subfigure[FC6+avg pooling]{\label{fig:size_FC6}%
\includegraphics[width=2.0in]{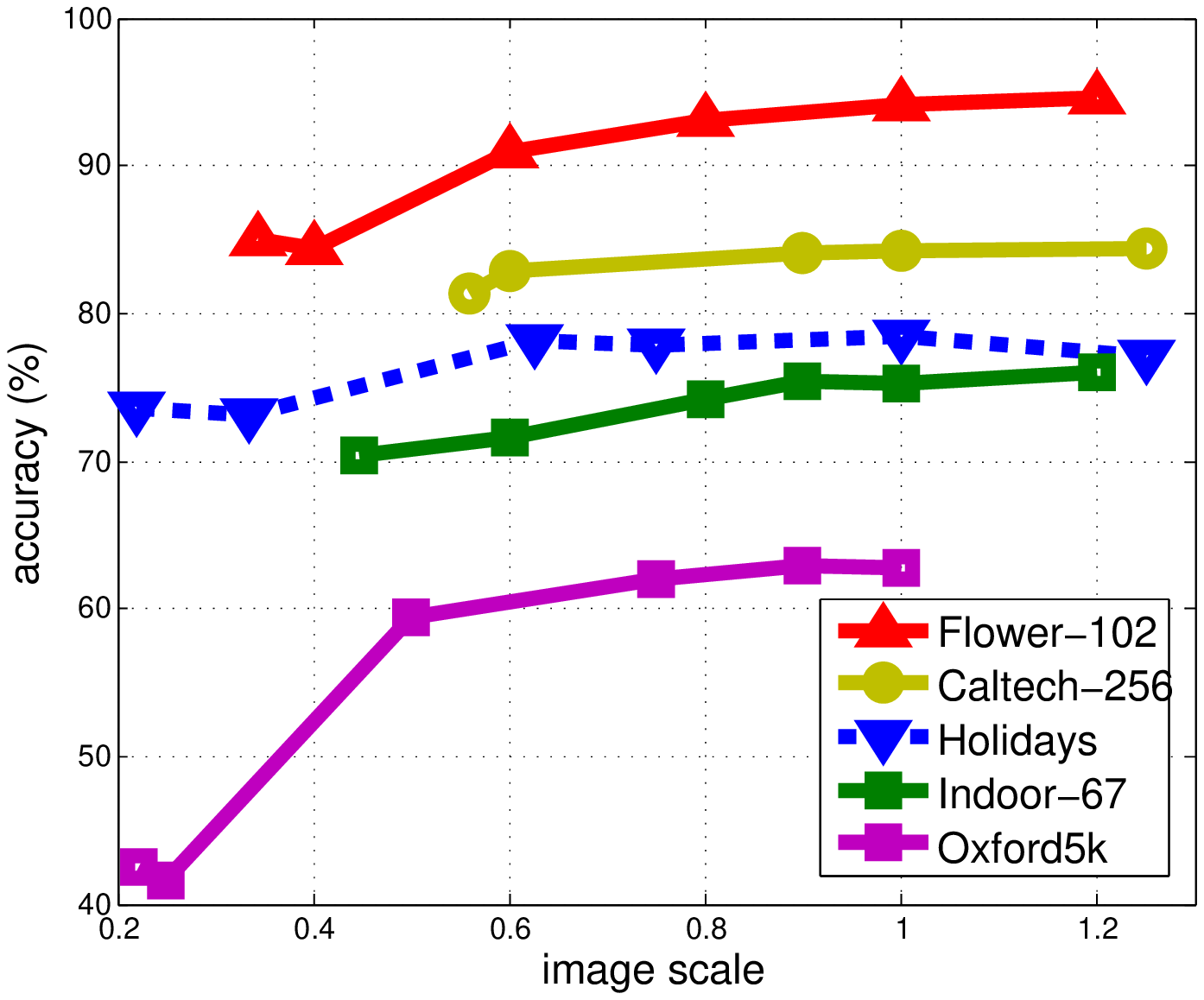}}
\hspace{0.18in}
 \subfigure[FC7+avg pooling]{\label{fig:size_FC7}%
\includegraphics[width=2.0in]{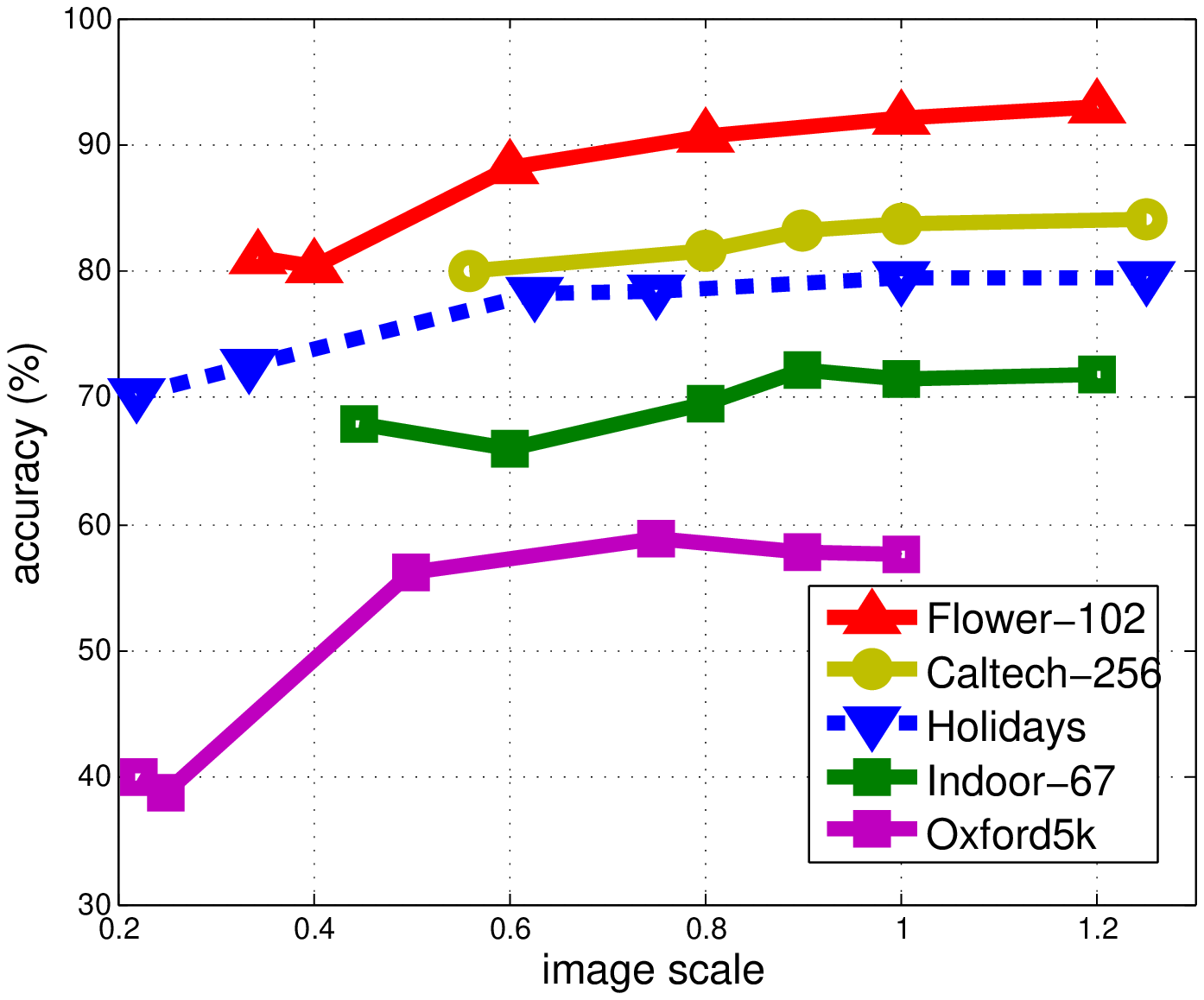}}
\caption{Impact of image size on five datasets. For scale = 1.0, all images are resized with equal long size determined in the training set, and aspect ratio preserved. The smallest scale corresponds to $224\times224$.}
\label{fig:image_size}
\end {figure*}
\begin{figure*}
  \centering
  \includegraphics[width=6.8in]{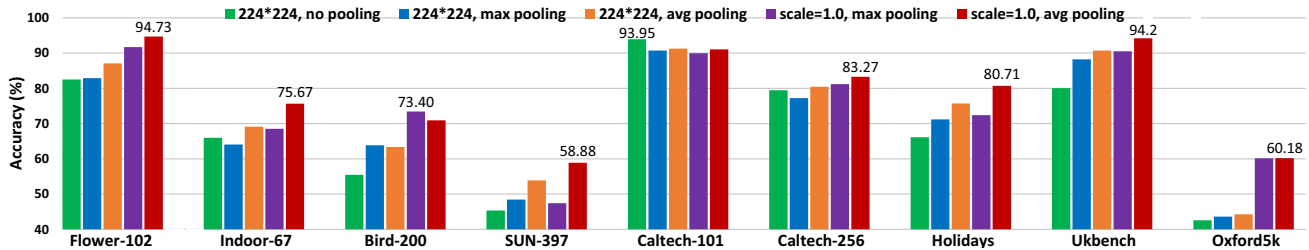}\\
  \caption{Comparison of average and max pooling of \textbf{Conv5 features} on 9 datasets. Two image sizes are shown, \emph{i.e.,} $224\times224$ and scale 1.0 (Section \ref{sec:role_size}). For each image size, max and avg pooling are compared.}\label{fig:avg_max}
\end{figure*}
\textbf{Impact of image sizes.} We then evaluate the role of image size. Various image scales are generated. Recall that scale 1.0 is so defined that images keep their aspect ratio and have the same long side as calculated in the training set (described in Section \ref{sec:role_size}).

By scaling images to various scales, we report the results on 5 datasets, \emph{i.e.,} Flower-102, Caltech-256, Holidays, Indoor-67, and Oxford5k, shown in Fig \ref{fig:image_size}. Note that for all the datasets shown, the smallest size corresponds to $224\times224$. It is clear from Fig. \ref{fig:image_size} that the increase in image scale consistently improves search and classification accuracy. This validates the assumption that using larger images and and keeping aspect ratio are beneficial for performance improvement. Moreover, we find that recognition accuracy remains stable when reaching scale 1.0. A scale larger than 1.0 would not incur much improvement because there is no information gain, and instead increase the memory consumption of GPU.

In addition to the above evidence, we also collect results on the ILSVRC'12 validation set \cite{ILSVRC15} which is not a transfer dataset. Using VGGNet, we average the output of the ``softmax'' layer, and sort the scores in descending order. Results in Table \ref{table:image_net} indicate that a larger input image yields lower classification error and that a scale larger than 1.0 does not help. In summary, the results are supporting our proposal that using properly large images and keeping aspect ration bring about decent improvement.

\setlength{\tabcolsep}{3.7 pt}
\begin{table}[!t]
\renewcommand{\arraystretch}{1.0}
\begin{center}
\begin{tabular}{|l|cccccc|}
\hline
\multirow{1}{*}{scale} &
 1.1 & 1.0 &0.95& 0.8& 0.6&0.45 \\
\hline
top-1 error & 25.91 & \bf25.45 &25.53&25.98& 30.41&34.12\\
top-5 error& 7.81  & \bf7.62&7.79&8.04&10.71& 13.17\\
\hline
\end{tabular}
\caption{Recognition error (\%) on ILSVRC'12 validation set. Scale = 0.45 corresponds to image resized to $224\times224$.}
\label{table:image_net}
\end{center}
\end{table}

 \textbf{Impact of pooling.} We demonstrate the effectiveness of pooling steps in Fig. \ref{fig:six_datasets} and Fig. \ref{fig:avg_max}. Two major conclusions are drawn. First, we observe from Fig. \ref{fig:six_datasets} that on all layers, pooling typically improves over directly using raw features. This is because raw features from the bottom layers are particularly sensitive to image variances such as translation or occlusion (FC features also have some sensitivity). Pooling reduces such effect by aggregating local activations into a global representation that is similar to the traditional Bag-of-Words model. Pooling also promotes computational efficiency by reducing the feature dimension.

Second, in Fig. \ref{fig:avg_max}, we provide a comparison between max and average pooling of the Conv5 feature in 9 testing benchmarks. We find that in most cases (8 out of 9 datasets), average pooling is superior to max pooling, except for the Bird-200 dataset. For Conv5 features, average pooling works well on datasets containing scenes or large objects.  But in the Bird-200 dataset (see Fig. \ref{fig:sample_images}), the target objects (birds) only takes up a very small region. As a result, the activation maps in the Conv5 layer should be sparse compared with the other datasets in which the objects (or scenes) are larger. In this situation, the effectiveness of average pooling is compromised by zeros in the maps.

\textbf{Impact of fusion.} The result of multi-layer feature fusion is summarized in Table \ref{table:various_approaches}. For image search, we combine multi-layer features by Query Adaptive Fusion (QAF) \cite{zheng2015query} on the score level. We observe consistent improvement on the three image search datasets. Specifically, we achieve +7.49\%, +0.07, and +11.12\% increase in mAP and N-S score, on Holidays, Ukbench, and Oxford5k datasets, respectively.

For image classification, by concatenating pooled features from the 7 layers, we observe consistent improvement in all 7 datasets, \emph{i.e.,} +2.95\%, +0.89\%, +2.75\%, +4.83\%, +0.07\%, +1.77\%, and +1.09\% on Bird, Flower, Indoor, SUN, Caltech-101, Caltech-256, and PASCAL VOC'07, respectively. The smallest improvement comes from Flower-102 and Caltech-101. On one hand, it is because the recognition accuracy on both datasets is already high enough (94.98\% and 92.24\% respectively). On the other, FC6 feature works well on Caltech-101, so the inclusion of lower-level features does not contribute much. Overall, these results indicate that features across multiple CNN layers are well-complementary with each other, so that their combination brings further improvement.

\subsection{Comparison with the State-of-the-arts}\label{sec:state_of_the_art}
We compare our results to state-of-the-arts in image search and classification. First, for fine-grained classification, \emph{i.e.,} Flower-102 and Bird-200, the comparison is presented in Table \ref{table:comparison_fine_grain}. Note that we do not use the bounding boxes and part annotations provided by the Bird-200 dataset. The recognition accuracy reported in this paper is \textbf{76.4\%} and \textbf{95.6\%} on Bird-200 and Flower-102, respectively, which exceed the state-of-the-art methods \cite{xiao2014application}\cite{qian2015fine} by +6.7\% and +6.1\%, respectively.

\setlength{\tabcolsep}{9.5pt}
\begin{table}[!t]
\renewcommand{\arraystretch}{1.0}
\begin{center}
\caption{Fine-grained recognition accuracy (\%).}
\begin{tabular}{|l|cc|}
\hline
\multirow{1}{*}{Methods} &
 Bird-200 & Flower-102 \\
\hline
\small CNNaug-SVM \cite{razavian2014cnn} & 61.8 & 86.8\\
\small ONE+SVM \cite{xie2015image} & 62.0  & 86.8\\
\small Two-level attention \cite{xiao2014application} & 69.7  & -\\
\small MsML+ \cite{qian2015fine} & 67.9  & 89.5\\
\small GMaxPooling \cite{murray2014generalized} &  33.3& 84.6\\
\small Color+SIFT \cite{khan2013discriminative} &  26.7 & 81.3\\
\hline
\hline
Ours (Alex) & 64.2 & 92.4 \\
Ours (VGG) & \textbf{76.4} & \textbf{95.6} \\
\hline
\end{tabular}
\label{table:comparison_fine_grain}
\end{center}
\end{table}

\setlength{\tabcolsep}{13pt}
\begin{table}[!t]
\renewcommand{\arraystretch}{1.0}
\begin{center}
\caption{Scene recognition accuracy (\%).}
\begin{tabular}{|l|cc|}
\hline
\multirow{1}{*}{Methods} &
 Indoor-67 & SUN-397 \\
\hline
\small CNNaug-SVM \cite{razavian2014cnn} & 69.0 & -\\
\small ONE+SVM \cite{xie2015image} & 70.1  & 54.9\\
\small MSOP \cite{gong2014multi} & 68.9  & 52.0 \\
\small Semantic FV \cite{dixit2015scene} &  72.9 & 54.4\\
\small PlacesCNN \cite{zhou2014learning} & 69.0 & 54.3\\
\hline
\hline
Ours (Alex)& 68.1& 52.0 \\
Ours (VGG)& \textbf{78.4} & \textbf{63.7} \\
\hline
\end{tabular}
\label{table:comparison_scene}
\end{center}
\end{table}
Second, for scene recognition, Table \ref{table:comparison_scene} summarizes the comparison with state-of-the-arts. On Indoor-67 and SUN-397 datasets, our system produces superior results: \textbf{78.4\%} and \textbf{63.7\%}, respectively. Comparing with ``ONE+SVM'' \cite{xie2015image}, our result on Indoor-67 and SUN-397 is higher by +5.5\% and +8.8\%, respectively.

For generic object classification on Caltech-101, -256, and PASCAL VOC'07 (Table \ref{table:generic}), our results still outperform the best recognition systems. Specifically, we report accuracy of \textbf{92.3\%}, \textbf{86.0\%}, and \textbf{83.7\%}, respectively.

Finally, for image search, we compare with state-of-the-art CNN methods in Table \ref{table:search}. This paper reports higher result: \textbf{mAP = 84.2\% and 71.3\%} on Holidays and Oxford5k, and \textbf{N-S = 3.75} on Ukbench. Note that the SIFT-based BoW model still outperforms CNN on Oxford5k due to the intensive illumination and viewpoint changes \cite{root_sift}. When integrating Graph Re-ranking \cite{zhang2012query}, we further improves the search accuracy to 89.3\%, 80.5\%, and 3.90, respectively.

\setlength{\tabcolsep}{7.6pt}
\begin{table}[!t]
\renewcommand{\arraystretch}{1.0}
\begin{center}
\caption{Generic object classification accuracy (\%).}
\begin{tabular}{|l|ccc|}
\hline
\multirow{1}{*}{Methods} &
 Cal-101 & Cal-256 &VOC'07\\
\hline
\small Zeiler-Fergus \cite{zeiler2014visualizing} & 86.5 & 74.2&-\\
\small Epitomic \cite{papandreou2014deep} & 87.8  & -&-\\
\small LLKNNC  \cite{liu2015novel} & -  & 75.5 &-\\
\small Image Codes \cite{kuang2015hardware} & 71.4 &  35.7& 52.9\\
\small FL+EN \cite{zhu2014submodular} & 83.2 & - &72.4\\
\hline
\hline
\small Ours (Alex) & {89.5} & {74.3} & 74.4\\
\small Ours (VGG) & \textbf{92.3} & \textbf{86.0} & \textbf{83.7}\\
\hline
\end{tabular}
\label{table:generic}
\end{center}
\end{table}

\setlength{\tabcolsep}{6.2pt}
\begin{table}[!t]
\renewcommand{\arraystretch}{1.0}
\begin{center}
\caption{Image search performance (mAP and N-S score).}
\begin{tabular}{|l|ccc|}
\hline
\multirow{1}{*}{Methods} &
 Holidays &Ukbench & Oxford5k\\
\hline
\small CNN basel. \cite{razavian2014cnn} & 84.3 & 3.64 & 68.0\\
\small CNN+PCA \cite{gong2014multi}& 80.2 & - &-\\
\small CNN+VLAD  \cite{ng2015exploiting}& 83.6 & - &59.3 \\
\small Neural Codes \cite{babenko2014neural}& 79.3& 3.56 & 54.5\\
\hline
\hline
Ours (VGG) & \textbf{84.2} & \textbf{3.75} & \bf71.3\\
  $\text{   }$  + re-ranking & \emph{89.3} & \emph{3.90} & \emph{80.5}\\
\hline
\end{tabular}
\label{table:search}
\end{center}
\end{table}

\section{Conclusion}\label{sec:conclusion}
In this paper, we share several findings with the community on the effective usage of CNN features during transfer. First, evidences accumulate that using larger images other than $224\times224$ yields superior accuracy. Second, the application of average/max pooling on the activation maps of intermediate CNN features consistently improves recognition performance over raw features. Specifically, we find that the pooled Conv5 feature produces superior or competitive performance to fully connected features.
Finally, the combination of features across multiple CNN layers further promotes recognition accuracy, and our system is capable of pushing the state-of-the-arts forward to a large margin.

Future work will focus on improving the discriminative power of bottom level CNN features. Possible strategies include injecting spatial constraints, multiple feature fusion, as well as effective encoding methods.

{\footnotesize
\bibliographystyle{ieee}
\bibliography{egbib}
}

\end{document}